\documentclass[letterpaper]{article}
\usepackage{style}
\usepackage{times}
\usepackage{helvet}
\usepackage{courier}
\usepackage[hyphens]{url}
\usepackage{graphicx}
\usepackage{natbib}
\usepackage{hyperref}
\usepackage{caption}
\usepackage{algorithm}
\usepackage{algorithmic}
\usepackage{multirow}
\usepackage{amsmath}
\usepackage{xcolor}       
\usepackage{subcaption}
\usepackage{newfloat}
\usepackage{listings}
\DeclareCaptionStyle{ruled}{labelfont=normalfont,labelsep=colon,strut=off}
\floatstyle{ruled}
\newfloat{listing}{tb}{lst}{}
\floatname{listing}{Listing}
\usepackage{tcolorbox}
\usepackage[framemethod=TikZ]{mdframed}
\newtcolorbox{prompt}{
  sharp corners,
  boxrule=0pt,
  boxsep=0pt,
  left=10pt,
  right=10pt,
  top=10pt,
  bottom=10pt,
  colback=gray!10,
}
\definecolor{delim}{RGB}{20,105,176}
\definecolor{string}{rgb}{0.64,0.08,0.08}
\definecolor{keycolor}{rgb}{0,0,1}
\definecolor{commentgreen}{rgb}{0,0.6,0}
\lstdefinelanguage{json}{
    numbers=left,
    numberstyle=\small,
    frame=single,
    rulecolor=\color{black},
    showspaces=false,
    showtabs=false,
    breaklines=true,
    breakatwhitespace=true,
    basicstyle=\ttfamily\small,
    upquote=true,
    morestring=[b]",
    belowskip=0.5em,
    stringstyle=\color{string},
    morecomment=[l]{:},
    commentstyle=\color{keycolor},
    literate=
     *{0}{{{\color{numb}0}}}{1}
      {1}{{{\color{numb}1}}}{1}
      {2}{{{\color{numb}2}}}{1}
      {3}{{{\color{numb}3}}}{1}
      {4}{{{\color{numb}4}}}{1}
      {5}{{{\color{numb}5}}}{1}
      {6}{{{\color{numb}6}}}{1}
      {7}{{{\color{numb}7}}}{1}
      {8}{{{\color{numb}8}}}{1}
      {9}{{{\color{numb}9}}}{1}
      {\{}{{{\color{delim}{\{}}}}{1}
      {\}}{{{\color{delim}{\}}}}}{1}
      {[}{{{\color{delim}{[}}}}{1}
      {]}{{{\color{delim}{]}}}}{1},
}
\newcommand{\methodname}{SwitchGPT}

\title{SwitchGPT: Adapting Large Language Models for Non-Text Outputs}
\author{
    Xinyu Wang\textsuperscript{\rm 1}, Bohan Zhuang\textsuperscript{\rm 2}, Qi Wu\textsuperscript{\rm 1}\thanks{Correspondence: \textit{Prof.} Qi Wu (qi.wu01@adelaide.edu.au)}
}
\affiliations{
    \textsuperscript{\rm 1} The University of Adelaide, Australia\\
    \textsuperscript{\rm 2} Monash Univeristy, Australia\\
}

\begin{document}

\maketitle

\begin{abstract}
Large Language Models (LLMs), primarily trained on text-based datasets, exhibit exceptional proficiencies in understanding and executing complex linguistic instructions via text outputs. However, they falter when requests to generate non-text ones. Concurrently, modality conversion models, such as text-to-image, despite generating high-quality images, suffer from a lack of extensive textual pretraining. As a result, these models are only capable of accommodating specific image descriptions rather than comprehending more complex instructions. To bridge this gap, we propose a novel approach, \methodname, from a modality conversion perspective that evolves a text-based LLM into a multi-modal one. We specifically employ a minimal dataset to instruct LLMs to recognize the intended output modality as directed by the instructions. Consequently, the adapted LLM can effectively summon various off-the-shelf modality conversion models from the model zoos to generate non-text responses. This circumvents the necessity for complicated pretraining that typically requires immense quantities of paired multi-modal data, while simultaneously inheriting the extensive knowledge of LLMs and the ability of high-quality generative models. To evaluate and compare the adapted multi-modal LLM with its traditional counterparts, we have constructed a multi-modal instruction benchmark that solicits diverse modality outputs. The experiment results reveal that, with minimal training, LLMs can be conveniently adapted to comprehend requests for non-text responses, thus achieving higher flexibility in multi-modal scenarios. Code and data will be made available at \href{https://github.com/xinke-wang/SwitchGPT}{https://github.com/xinke-wang/\methodname}.
\end{abstract}

\section{Introduction}

\begin{figure}[t!]
    \centering
    \includegraphics[width=\linewidth]{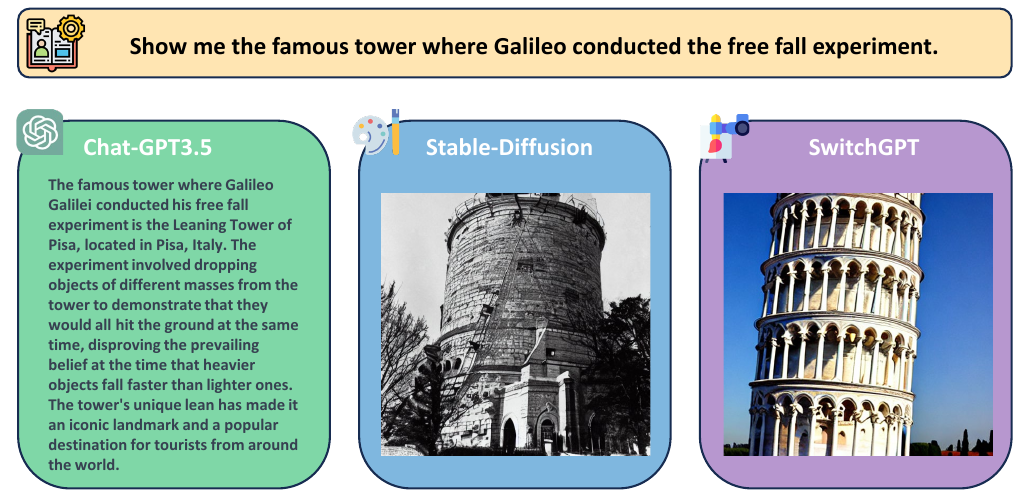}
    \caption{Given an instruction expecting a non-text response, text-based LLMs like ChatGPT~\cite{openai2023chatgpt} are constrained to providing text responses, while popular text-to-image models such as Stable Diffusion~\cite{rombach2022high} generate imagery based on direct description. In contrast, our proposed \methodname\
    comprehensively interprets the underlying intent of the instruction, accurately producing a more appropriate response. 
    }
    \label{fig:ChatGPT-StableDiffusion-Ours}
\end{figure}

The emergence of Large Language Models (LLMs)~\cite{brown2020language, zhang2022opt, touvron2023llama, chowdhery2022palm} that are capable of understanding and executing complex linguistic tasks has been a remarkable advancement in recent years. Their prowess in comprehending, processing, and generating text-based responses has paved the way for groundbreaking applications. This includes fields such as natural language understanding~\cite{wang2018glue}, automated question-answering systems~\cite{lin2021truthfulqa}, and conversational AI assistants~\cite{openai2023chatgpt}. Predominantly, these models are trained on extensive crowdsourced text-based datasets. Such datasets encompass a myriad of topics and languages, capturing the vast expanse of human knowledge~\cite{raffel2020exploring, radford2019language}. Nonetheless, due to their text-centric training, traditional LLMs primarily operate with textual inputs and outputs, resulting in unsatisfactory responses when tasked with generating non-textual outputs (see Figure~\ref{fig:ChatGPT-StableDiffusion-Ours}). Despite not being directly exposed to non-textual data such as images and speech during their training, recent research~\cite{li2023blip, wang2023makes, li2023lmeye, su2023pandagpt, zhu2023minigpt} has indicated that LLMs possess a profound potential to understand such non-textual data. This revelation opens the door to evolving pure text-based LLMs into a multi-modal paradigm. For example, Mini-GPT4~\cite{zhu2023minigpt} trains a linear layer that connects BLIP-2~\cite{li2023blip} with Vicuna~\cite{chiang2023vicuna}, demonstrating the possibility that LLMs can understand image inputs. However, 
studies on enabling LLMs to produce non-textual outputs remain relatively limited, restricting the LLM's interactive capabilities in multimodal scenarios.

Unlike traditional LLMs, which are primarily designed for unimodal interaction, specifically text-to-text communication, modality conversion models are adept at handling data across different modalities. For example, image captioning~\cite{vinyals2015show, hossain2019comprehensive} exemplifies the route of image$\rightarrow$text, whereas text-conditioned image generation~\cite{goodfellow2014generative, ramesh2021zero} illustrates the transition from text$\rightarrow$image. These models signal a significant advancement in producing realistic samples across diverse data types. Yet, their training chiefly relies on paired data, such as image-text pairs. The available volume of such paired data is substantially smaller than single modality data, with comparisons often being in the ballpark of hundreds of billion tokens (pure text) \textit{v.s.} mere hundreds of million pairs (image-text). As a result, modal conversion models often lack the richness of knowledge and depth of understanding exhibited by LLMs. This limitation also means they usually struggle with comprehending and executing more complex instructions that rely on acquired knowledge and common sense. (see Figure~\ref{fig:ChatGPT-StableDiffusion-Ours}).

Given the aforementioned strengths and limitations of both LLMs and modality conversion models, an intriguing proposition emerges: \textit{Can we amalgamate the profound knowledge and understanding capabilities of LLMs with the modality conversion models?} Hence, the integrated models could potentially interpret more intricate instructions and deliver outputs in various modalities. A natural idea that emerges from this conundrum is to position the LLM as a coordinator to orchestrate and utilize modality conversion models. Some recent works~\cite{schick2023toolformer, shen2023hugginggpt, lian2023llm} have demonstrated the immense potential of leveraging LLMs as a controller, showcasing their superior orchestration abilities to plan and execute fine-grained instructions with external tools. For instance, HuggingGPT~\cite{shen2023hugginggpt} devised a workflow where ChatGPT is used as a controller to invoke HuggingFace's open-sourced models, thereby accomplishing sophisticated AI tasks. However, it suffers from a few shortcomings. \textbf{Inefficiency:} HuggingGPT operates as an online model, heavily leaning on OpenAI's ChatGPT API. This dependency necessitates frequent invocations of ChatGPT for functions such as Task Planning and Model Selection, substantially inflating both the operational cost and latency. \textbf{Instability:} HuggingGPT uses ChatGPT as a black box without any tuning, and the LLM's outputs are uncontrolled and do not always return the desired results, leading to exceptions in the workflow. Another example is LLM-Grounded Diffusion~\cite{lian2023llm}, which harnesses the reasoning capabilities of LLMs to produce scene layouts in the form of bounding boxes based on the input instructions. As a result, it offers more meticulous control over the positioning and layout of objects with the generated image. However, the \textbf{inflexibility} of such methods confines them to a singular modality conversion pathway, and they often falter when interpreting indirect requests. The above shortcomings highlight the challenges in current integration efforts and the full potential of combining LLMs with modality conversion models.

In this paper, we endeavor to address the above challenges and present an instruction-tuned LLM to unlock its ability to generate non-text outputs, with the help of several modality conversion models. The primary contributions of this paper are threefold:

\begin{itemize}
    \item We present a Modality-aligned Instruction Tuning that efficiently enables LLMs to discern the intended output modality as dictated by the instructions. Thus, LLMs are empowered to summon appropriate modality conversion models for non-text responses while fully retaining their original reasoning capabilities. With this technique, any LLM can be easily adapted to produce non-text outputs with minimal training.
    \item We introduce a new evaluation set that includes thousands of instructions targeting text, image, and audio outputs, to assess LLMs' abilities in handling multi-modal output requests, enabling a better understanding of performance variations across different LLMs in multi-modal scenarios.
    \item We conduct comprehensive experiments to validate our approach against state-of-the-art open-source LLMs as well as OpenAI's ChatGPT API in multi-modal scenarios. The experimental results indicate that our models tuned with modality-aligned instructions are able to retain their original reasoning abilities while consistently generating accurate and appropriate non-text outputs.
\end{itemize}

\section{Related Work}

\textbf{LLM as Controller.}
The AI community has witnessed a transformative surge in Natural Language Processing advancements over recent years, largely driven by the emergence of LLMs, such as GPT~\cite{brown2020language}, OPT~\cite{zhang2022opt}, PaLM~\cite{chowdhery2022palm}, Bloom~\cite{scao2022bloom}, and LLaMa~\cite{touvron2023llama}. The potential of LLMs extends beyond their immediate function of text generation. With vast knowledge bases and intricate reasoning capabilities, LLMs closely emulate human-like comprehension. With these capabilities, a promising direction has emerged, which treats the LLM as a controller. Instead of using LLMs merely as standalone entities, they can serve as a coordinator between external tools~\cite{shen2023hugginggpt, schick2023toolformer, li2023api, qin2023toolllm} or even manipulate robotic systems~\cite{driess2023palm, mai2023llm}. For example, HuggingGPT~\cite{shen2023hugginggpt} employs ChatGPT as a controller to manage a list of open-sourced models in HuggingFace's Hub to solve AI tasks. However, these methods often treat LLMs as a black box, utilizing custom prompts for specific invocation rules, leading to notably unstable outputs. Contrary to solely depending on pre-trained LLMs for generating control commands, our method fine-tunes the LLM to yield structured responses, thereby ensuring more consistent and reliable outcomes.

\begin{figure*}[t!]
    \centering
    \includegraphics[width=0.7\linewidth]{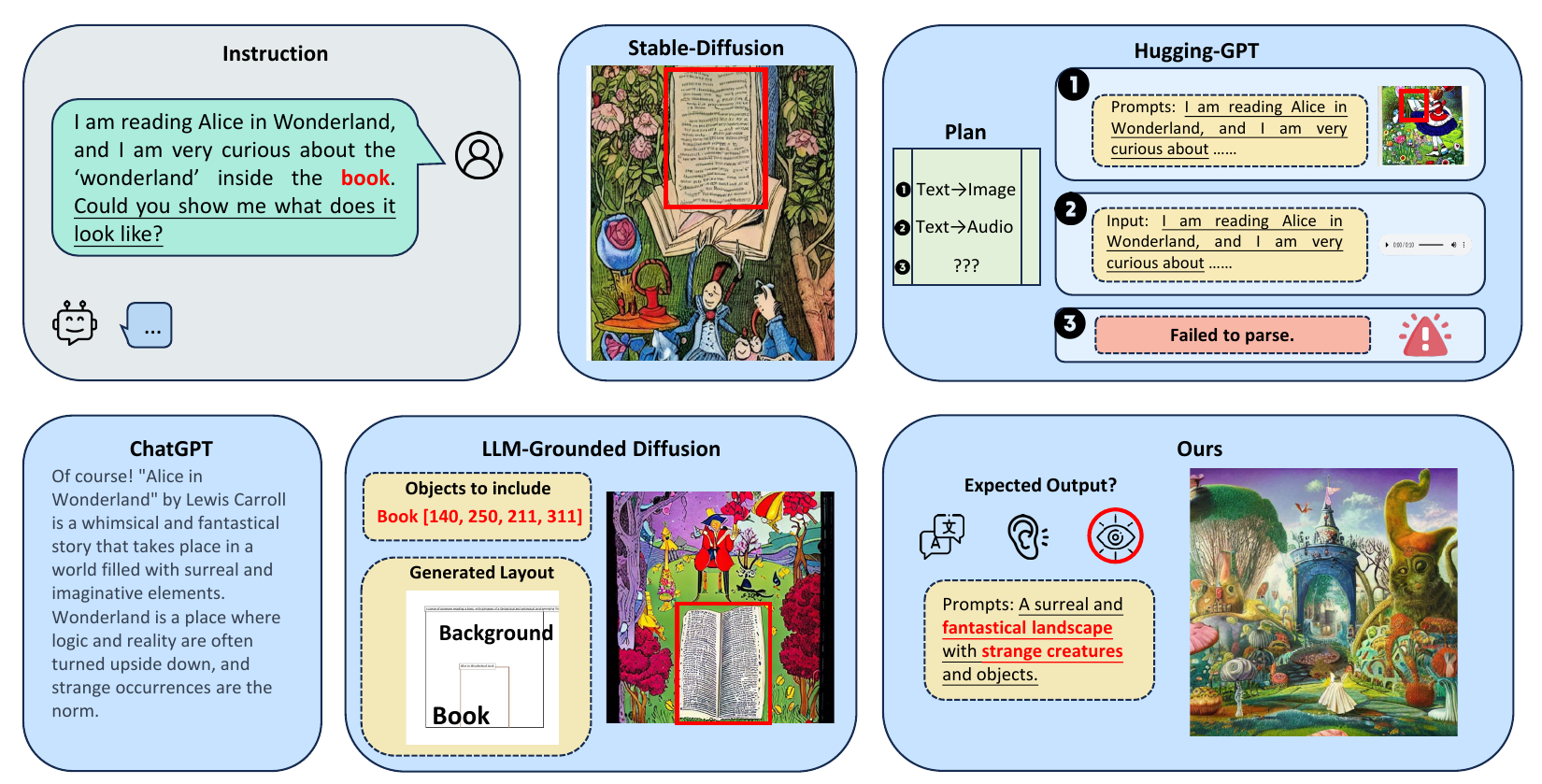}
    \caption{Comparison of responses to the user's instruction by different models. While traditional text-to-image models like Stable Diffusion often generate images based on superficial keywords, they might miss the underlying intent of the instruction. Hugging-GPT can produce unstable results, for instance, it outputs three responses in this case. LLM-Grounded Diffusion, though adept at controlling the layout, still falls short in grasping the deeper nuances behind the user's request, as evidenced by the unwarranted inclusion of a `book' in its image. In contrast, our proposed approach not only captures the true essence of the instruction but also aligns it with the desired input for the modality conversion model, resulting in a more faithful visual representation. (The figure is best viewed zoomed in.)}
    \label{fig:Comparison-HuggingGPT-LLMDifusion}
\end{figure*}

\noindent\textbf{LLM Finetuning.}
Given that LLMs typically possess tens to hundreds of billions of parameters and necessitate training on vast datasets, the barriers to their training and fine-tuning are significantly raised, both in terms of computational resources and data collection. Recently, many efforts have delved into more resource-efficient methods of LLM finetuning. From a data perspective, instruction tuning~\cite{wang2022self, liu2023visual, alpaca} leverages high-performing, large-scale LLMs such as GPT-3.5 or GPT-4 to generate vast amounts of instructions and responses, which are then used to fine-tune relatively smaller LLMs or even those large models themselves. From the view of model parameters, Parameter Efficient Fine-Tuning (PEFT) explores the use of post-training quantization~\cite{dettmers2022llm, frantar2022gptq} or freezing LLM parameters to train an adapter~\cite{hu2021lora, dettmers2023qlora}, with the goal of reducing computational overhead. Based on these techniques, low-cost, customizable training of LLMs has become feasible, for example, Alpaca~\cite{alpaca} and Vicuna~\cite{chiang2023vicuna} significantly improve the performance of Llama~\cite{touvron2023llama} by conducting an instruction tuning. LLaVA~\cite{liu2023visual} presents visual instruction tuning, enabling LLMs to understand image contents by tuning on language-image paired instructions. However, these efforts either focus on enhancing the LLM's reasoning performance or on its image$\rightarrow$text understanding abilities. In contrast, our approach delves deeper into understanding the intention behind input instructions, enabling the LLM to decide the most fitting output modality. This unique capability is achieved using our proposed modality-aligned instruction tuning, which not only retains the text$\rightarrow$text reasoning prowess of the LLMs but also enables them to produce non-text outputs, including text$\rightarrow$image and text$\rightarrow$speech.

\begin{figure*}[t!]
    \centering
    \includegraphics[width=\linewidth]{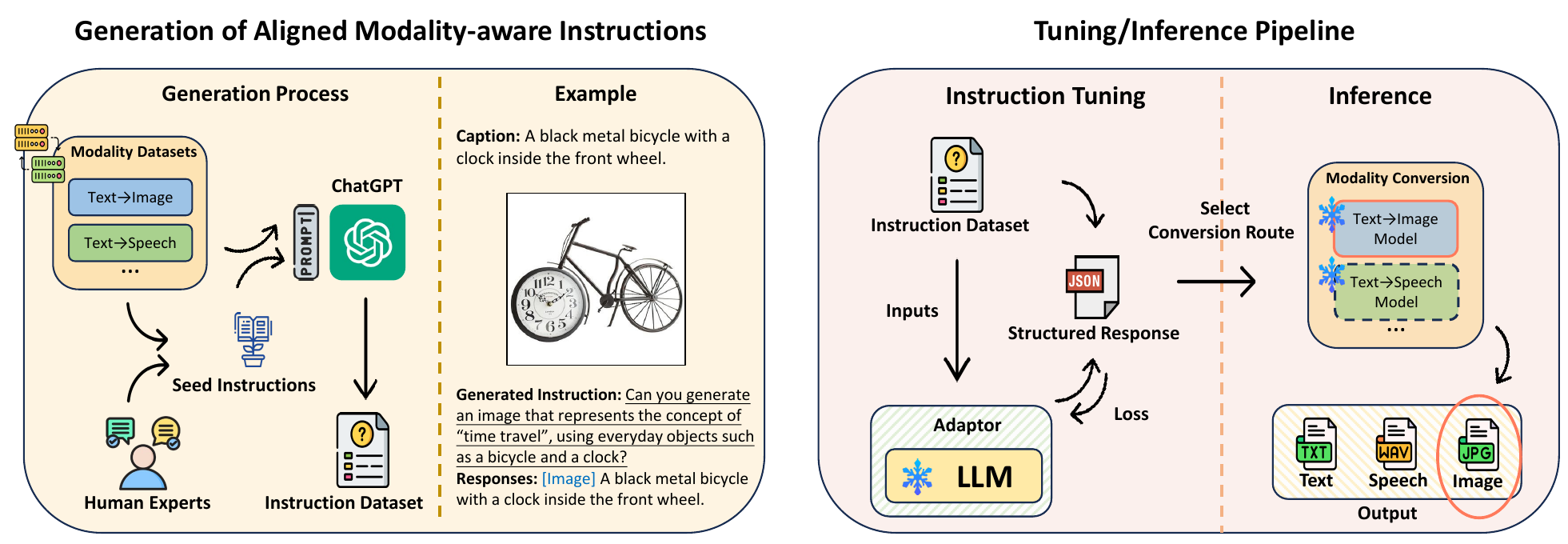}
    \caption{\textbf{Left:} Flowchart illustrating the process of generating modality-aligned instructions. By embedding the text from modality datasets into prompts, our method ensures alignment between ChatGPT-generated instructions and the modality conversion models. \textbf{Right:} Pipeline of instruction tuning and inference. During training, the parameters of the LLM are frozen, and an adapter is trained on the generated modality-aligned instructions dataset. At the inference stage, the structured response of LLM is parsed and used to select the appropriate modality conversion route to produce the final outputs.}
    \label{fig:pipeline}
\end{figure*}

\noindent\textbf{Modality Conversion Models.}
Multi-modal applications have been a long-standing research topic in the AI community. The unique capability of these models to comprehend and translate between different data representations facilitates conversions between various modalities, such as image captioning~\cite{vinyals2015show, hossain2019comprehensive} (image$\rightarrow$text), text-conditioned image generation~\cite{goodfellow2014generative, ramesh2021zero} (text$\rightarrow$image), speech recognition~\cite{ao2021speecht5} (audio$\rightarrow$text), or even multiple modality conversion (text/image$\rightarrow$text/image) such as Uni-Diffuser~\cite{bao2023one}. We specifically focus on text$\rightarrow$image and text$\rightarrow$speech conversion in this paper. Given a text description, a text$\rightarrow$image model aims at synthesizing an image that accurately depicts the content described in the text. Early methods were mostly based on GAN~\cite{xu2018attngan, goodfellow2014generative} and VQ-VAE~\cite{ramesh2021zero}, while recently, due to their stability and improved generation quality, diffusion-based~\cite{rombach2022high} approaches have become increasingly popular. For text$\rightarrow$speech, the goal is to transform a piece of text content to corresponding audio. While these models can convert text into other modalities, they struggle to understand complex instructions, making it difficult to apply them to advanced interactive features, such as AI assistants. To tackle these challenges, our method synergizes LLMs with modality conversion models, thereby aligning the advanced reasoning abilities inherent to LLMs with the conversion capabilities of multi-modal models.

\section{\methodname}

\subsection{Preliminary}
\textit{Why multi-modal output matter?}
While the text remains the predominant medium for human communication, images, and sounds often assume indispensable roles in various contexts. For instance, a photograph capturing a golden sunset over a tranquil beach can elicit emotions that mere words may find challenging to express. Similarly, visually impaired individuals heavily depend on auditory cues and descriptions to comprehend their surroundings. Therefore, the value of non-text outputs cannot be understated. 

However, there is still limited research on adapting LLMs for non-text outputs. One primary reason for the limited research in this area is the prohibitive cost associated with multi-modal LLM pre-training. A pioneering work Emu~\cite{sun2023generative} aims to bridge this gap by introducing a vision-language pre-trained LLM. Emu is pre-trained on massive datasets, comprising billions of image-text pairs and millions of video-subtitle pairs, using 128 A100 GPUs. This enables Emu to not only be capable of accepting image/text inputs but also generate textual and visual outputs. Even though Emu has achieved impressive performance on various tasks, its adaptability comes at a high cost. For example, introducing a new modality such as audio necessitates the collection of corresponding paired data and complete retraining.

Given the flourishing developments in recent years, the AI community has amassed countless open-source models for diverse tasks. Integrating these models into LLMs offers another solution for adapting them to non-text outputs. This not only conserves computational resources and reduces carbon emissions but also provides a cost-effective and smooth transition to newer models without the need for extensive retraining. This paper falls into this category. Existing methods such as HuggingGPT~\cite{shen2023hugginggpt} can invoke external models to generate non-text responses beyond the capabilities of traditional LLMs like ChatGPT. However, its results are unstable. For example, Figure~\ref{fig:Comparison-HuggingGPT-LLMDifusion} depicts that HuggingGPT may produce an audio response when an image output is expected. Moreover, it inputs the full instruction as a prompt to Stable Diffusion, leading to suboptimal results. The reasons leading to the above issues can be summarized as follows:

\begin{itemize}
    \item Relying solely on designing prompts to utilize the zero-shot capability of LLM for invoking external models can introduce ambiguity, leading to unstable outputs.
    \item The misalignment between LLMs outputs and external model inputs often results in subpar performance.
\end{itemize}

\subsection{Modality-aligned Instruction Generation}
\label{sec:instruction-generation}

To solve the aforementioned issues, we introduce Modality-aligned Instruction Tuning (MaIT). The primary purpose of MaIT is twofold. First, MaIT aims to cheaply tune the LLM $L$ to understand and interpret the expected output modality $t$ from a given instruction $I$. Formally, this can be written as:

\begin{equation}
    L(I) = (r) \overset{\text{MaIT}}{\longrightarrow} L'(I) = (r, t),
\label{eq:initial-purpose}
\end{equation}

\noindent where $L'$ is the adapted LLM after instruction tuning, and $r$ is the output response. For example, consider a simple instruction ``What is the answer to the following equation 1+1=?". While $L$ might respond to `2', $L'$ is expected to produce (`2', `text'), with `text' being a flag denoting the desired output modality. This modality type flag, $t$, informs the LLM about \textit{when} to invoke \textit{which} modality conversion model. For instance, if $t$ is `image', the LLM knows to use text$\rightarrow$image conversion model rather than simply giving the response $r$. However, recognizing when and which model to invoke is not the sole requirement. It is equally crucial to guide the LLM on \textit{how} to use the modality conversion model. Without this guidance, the LLM might misinterpret the instruction or use a misaligned response as input when interfacing with the modality conversion model. This can be seen in the failure example of HuggingGPT depicted in Figure~\ref{fig:Comparison-HuggingGPT-LLMDifusion}, which results in less-than-ideal outcomes. As such, a secondary goal of MaIT is to ensure the LLM's outputs align seamlessly with the inputs of the modality conversion model. This is built upon the fact that for a single route of modality conversion, such as text$\rightarrow$image, different models often share almost the same training dataset. Therefore, instead of aligning the LLM with a particular model, it is more effective to align it with the training samples themselves. In other words, the objective is to minimize the distribution gap between the LLM's outputs and the textual descriptions in the modal conversion task training dataset. We can rewrite Equation~\ref{eq:initial-purpose} to represent this purpose as follows:

\begin{equation}
L(I) = (r) \underset{\min \Delta(D_{r'}, D_{\text{text}\rightarrow t})}{\overset{\text{MaIT}}{\longrightarrow}} L'(I) = (r', t),
\label{eq:rewrite-purpose}
\end{equation}

\noindent where $D_{r'}$ and $D_{\text{text}\rightarrow t}$ respectively represent the distribution of the output $r'$
by the adapted $L'$, and the text description distribution of the training dataset for the text$\rightarrow t$ task. To achieve this, we directly use the training data from the modality conversion tasks, such as the caption of images, to construct the response.

According to Equation~\ref{eq:rewrite-purpose}, we are able to generate modality-aligned instructions. Specifically, for the modality type $t$, we consider the three most common modalities in this paper, \emph{i.e.}, text, image, and speech. Following the approaches of previous studies~\cite{alpaca}, we initiate the process by designing seed instructions. Each record comprises three components: the instruction, the anticipated output modality, and the response. For example, a record might look like this: \{``instruction": ``How do you pronounce the name of the fast food brand with a yellow golden arch logo?", ``response": \{``type": ``speech", ``McDonald's"\}\}. Notably, with an abundance of open-source text-to-text instruction datasets available, we concentrated our efforts on devising seed instructions specifically for the text$\rightarrow$image and text$\rightarrow$speech modalities. To generate instructions in larger quantities, we employed OpenAI's ChatGPT API. This involved integrating descriptions from the modality conversion task's training data into the ChatGPT prompt to produce modality-aligned instructions. An example of this process is illustrated in Figure~\ref{fig:pipeline}. Using a picture caption like ``A black metal bicycle with a clock inside the front wheel", we prompt ChatGPT to formulate a suitable instruction that solicits the generation of such an image. In this case, the ChatGPT comes up with an intriguing instruction ``Can you generate an image that represents the concepts of `time travel', using everyday objects such as a bicycle and a clock?". This showcases ChatGPT's strong capability in producing diverse instructions. Additionally, the original caption, combined with the specified modality type, is employed to construct the ground-truth response. Specifically, we use the LAION-aesthetic~\cite{schuhmann2022laion} and LibriTTS~\cite{zen2019libritts} as referenced modality datasets to sample image captions or speech contents. In this manner, we synchronize the output of LLM with the input of the modality conversion model. Importantly, since we exclusively used textual descriptions of images for instruction creation, it is only necessary to conduct text-to-text tuning on the LLM. This avoids potential computational overhead from multi-modal fine-tuning. We will provide more details about the generated instructions in the appendix.

\begin{table*}[t]
\centering
\begin{tabular}{l|c|cc|c|c}
\hline
\multirow{2}{*}{Method} & \multirow{2}{*}{Modality Acc. (\%) $\uparrow$} & \multicolumn{2}{c|}{Vision} & Language & Speech \\
 & & CLIP $\uparrow$ & FID $\downarrow$ & QA $\uparrow$ & BLEU $\uparrow$ \\\hline
OPT-2.7B~\cite{zhang2022opt} & 68.4 & 16.6 & 131.3 & 0.53 & 0.07 \\
OPT-6.7B~\cite{zhang2022opt} & 16.2 & 6.4 & 277.3 & 0.66 & 0.00 \\
Llama-7B~\cite{touvron2023llama} & 66.8 & 17.9 & 88.6 & 0.55 & 0.07 \\
Llama2-7B~\cite{touvron2023llama2} & 76.2 & 18.7 & 86.1 & 0.69 & 0.08 \\
Alpaca-7B~\cite{alpaca} & 72.3 & 17.2 & 94.7 & 0.58 & 0.07 \\
Vicuna-7B~\cite{chiang2023vicuna} & 73.1 & 17.0 & 95.0 & 0.59 & 0.07 \\
GPT-3.5-turbo~\cite{openai2023chatgpt} & \textbf{88.1} & 21.4 & 84.1 & \textbf{0.75} & \textbf{0.17} \\
HuggingGPT~\cite{shen2023hugginggpt} & 81.7 & 19.9 & 85.2 & 0.62 & 0.10 \\
\methodname-7B (Ours) & 86.9 & \textbf{22.6} & \textbf{82.4} & 0.67 & 0.16 \\\hline
\end{tabular}
\caption{Comparison of the performance between the proposed \methodname\ and state-of-the-art models.}
\label{table:comparison-with-sota}
\end{table*}

\subsection{Training and Inference}
Once the modality-aligned instruction dataset has been generated, we integrate it with the existing text-only instructions. This results in the training set comprising three routes of instructions, \emph{i.e.}, text$\rightarrow$text, text$\rightarrow$image, and text$\rightarrow$speech. In line with previous research, we maintain a comparable number of instructions for training, totaling approximately 52k, with each route accounting for roughly one-third of this total. While conducting instruction-tuning, it is essential to preserve the original reasoning and generation capabilities of the LLM. Therefore, for Equation~\ref{eq:rewrite-purpose}, when $t$ is text, it is desired that the distribution of responses $r$ and $r'$ remain as similar as possible:

\begin{equation}
\left\{
\begin{aligned}
&\min\Delta(D_{r'}, D_{r})  &&, t = \text{text}  \\
&\min\Delta(D_{r'}, D_{\text{text}\rightarrow t}) &&, t \neq \text{text} \\
\end{aligned}
\right.
\end{equation}

\noindent This consistency is attained by incorporating the text$\rightarrow$text instructions during the instruction-tuning phase. Furthermore, the parameters of the LLM are kept constant. Only a Low-Rank adapter (LoRA)~\cite{hu2021lora} is fine-tuned, which not only safeguards the pre-trained weights but also significantly reduces computational costs. 

As shown in Figure~\ref{fig:Comparison-HuggingGPT-LLMDifusion}, a challenge of the current method such as HuggingGPT~\cite{shen2023hugginggpt} is the unpredictable format of the output. Such variability can lead to inconsistencies during subsequent parsing stages, often resulting in exceptions. To address this issue, we encode all responses into a structured JSON format, ensuring that the LLM produces outputs with a consistent structure. Due to its extensive pre-training, LLM is already familiar with the JSON formatting rules. Therefore, training with a modest amount of data can effectively guide it to generate appropriately formatted outputs. As shown in the right block of Figure~\ref{fig:pipeline}, the instructional data consists solely of text, so the LLM does not require interaction with any modality conversion models during the training phase. When moving to the inference stage, the LLM outputs a structured JSON format response for the given instruction. This determines \textit{when} to choose \textit{which} modality conversion model and \textit{how} to use it. In this manner, all modality conversion models remain static without any fine-tuning.

\section{Experiments}

\subsection{Implementation}

In implementing the proposed pipeline, we used the Llama-2~\cite{touvron2023llama2} as our foundation LLM. For modality conversion, StableDiffusion-v1-5~\cite{rombach2022high} and SpeechT5~\cite{ao2021speecht5} models were employed. The parameters of the LLM were loaded using \texttt{Int8}~\cite{dettmers2022llm} and subsequently frozen. With the modality-aligned instructions generated by GPT-3.5-turbo, we trained a LoRA adapter~\cite{hu2021lora} upon the frozen LLM. Our training was executed on four Nvidia A100 (40GB) GPUs (however, training on a single GPU is possible), utilizing a per-device batch size of 4 and gradient accumulation steps of 8. Optimization of the model was carried out using the AdamW~\cite{loshchilov2017decoupled} for 3 epochs, with a consistent learning rate $3^{-4}$. The entire training procedure was efficiently completed in approximately 3 hours. 

\subsection{Validation Set and Metrics}

Evaluating LLM performance objectively is crucial for contrasting different methods. Current benchmarks, such as The Open LLM Leaderboard~\cite{open-llm-leaderboard}, are primarily tailored to assess text$\rightarrow$text reasoning and generation capabilities. In contrast, this paper focuses on enabling the LLM to produce non-text outputs. Given this discrepancy, existing evaluation standards are not suited for a comprehensive assessment of our model. To address this gap, we introduce a new validation set (which will be released) specifically designed to facilitate the assessment of multi-modal output LLMs. This set contains 2,400 instructions, each demanding an output in one of three modalities: text, image, or speech. For the text$\rightarrow$text instructions, we sample questions from established benchmarks, including Truthful-QA benchmark~\cite{lin2021truthfulqa} and MMLU~\cite{hendrycks2020measuring}, applying a variant multiple-choice metric. The text$\rightarrow$image instructions are bifurcated into two categories: 200 intricate instructions crafted by humans, and 600 instructions chosen from proposals generated by ChatGPT-4 using the COCO-caption~\cite{chen2015microsoft} as a reference. The performance of LLM on this task is evaluated using both CLIP~\cite{radford2021learning} and FID~\cite{heusel2017gans} scores. Regarding the text$\rightarrow$speech tasks, a similar strategy as of the text$\rightarrow$image is used to construct the instructions, while Libritts~\cite{zen2019libritts} is employed as reference. As the text$\rightarrow$speech model translates input text verbatim, the audio quality remains unaffected by the LLM's output. We thus employ the BLEU~\cite{papineni2002bleu} score to gauge the similarity between the LLM's output and the referenced contents. Additionally, to ascertain if the model can discern the desired output modality from the instruction, we evaluate its performance using classification accuracy, and only those predictions that match the ground-truth modality will be further assessed for their vision, language, or speech scores. Notably, since the default output modality of LLM is text, these outputs will not contribute to classification accuracy. We explain more details about evaluation metrics in the appendix.

\begin{table}[t!]
\centering
\begin{tabular}{ccc}
\hline
Method & CLIP & FID \\ \hline
Stable Diffusion v1-5 & 15.2 & 140.3 \\
LLM-grounded Diffusion & 18.9 & 85.9 \\
\methodname-7B (Ours) & \textbf{22.6} & \textbf{82.4} \\ \hline
\end{tabular}
\caption{Comparison of the performance between Stable Diffusion v1-5, LLM-grounded Diffusion, and our method for text$\rightarrow$image instructions.}
\label{table:comparision-with-stable-diffusion}
\end{table}

\begin{figure*}[t!]
    \centering
    \includegraphics[width=\linewidth]{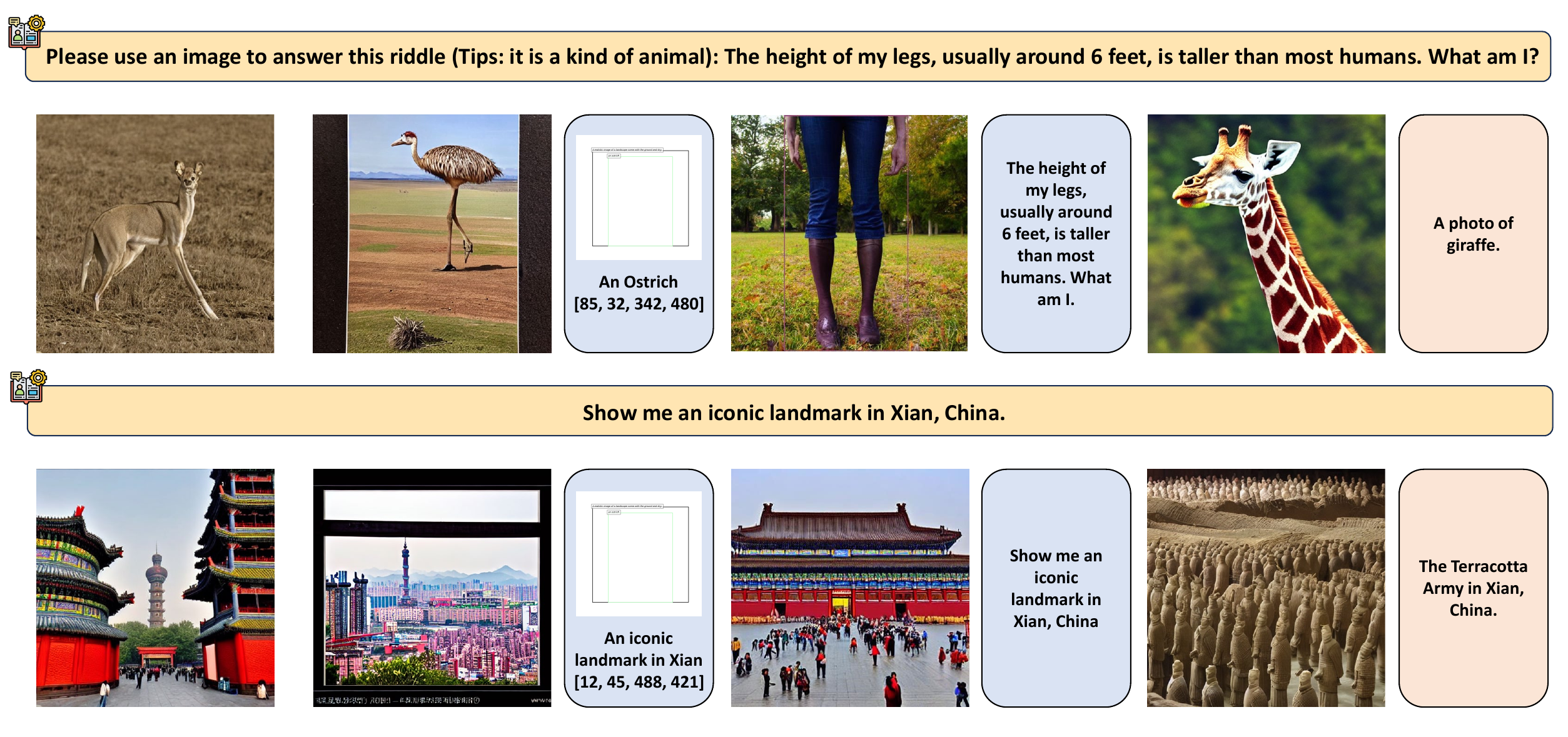}
    \caption{Qualitative results. From left to right: images produced by Stable Diffusion, LLM-grounded Diffusion, HuggingGPT, and the proposed \methodname. The right side of each image showcases the intermediate artifacts such as the layout and prompts (Stable Diffusion uses instruction as prompt directly) generated by the corresponding methods.}
    \label{fig:Qualitative-results}
\end{figure*}

\subsection{Quantitative Results}

To assess the efficacy of the proposed \methodname, we compared them with a range of state-of-the-art techniques. These include pretrained LLMs like OPT~\cite{zhang2022opt} and Llama~\cite{touvron2023llama, touvron2023llama2}; instruction-tuned LLMs such as Alpaca~\cite{alpaca} and Vicuna~\cite{chiang2023vicuna}; and the commercial API, GPT-3.5-turbo. Additionally, we considered methods that employ LLMs as controllers, such as HuggingGPT~\cite{shen2023hugginggpt} and LLM-grounded Diffusion~\cite{lian2023llm}. Given that many text-based LLMs are not inherently designed to produce non-text responses, a few-shot assessment approach was adopted. Specifically, four instructions and corresponding responses are given as examples, prompting these methods to generate responses to new inputs (please refer to supplementary for more details). Notably, all of the LLMs share identical modality conversion models to produce non-text responses.

Table~\ref{table:comparison-with-sota} demonstrates the performance of the proposed method in comparison with existing techniques. For modality accuracy, our method achieves comparable performance with GPT-3.5-turbo and surpasses all other techniques. This highlights its capability to understand the expected modality of the instruction. Interestingly, the OPT-6.7B model lags considerably in accuracy when compared to all other methods, even its own 2.7B counterpart. We observed that it predominantly provides text output for most instructions. This discrepancy might be attributed to its heightened sensitivity to specific prompts and examples. The CLIP score indicates the congruence of the generated images with their corresponding descriptions. As all LLM models utilize the same text$\rightarrow$image foundation model, the score can be used to measure the quality of the prompt generated by the LLM. the score serves as an indicator of the prompt quality produced by the LLM. Our model consistently excels in this metric, highlighting the effectiveness of the modality-alignment instruction tuning. Regarding language scoring, our method achieves an accuracy comparable to the original Llama2-7B. This suggests that the foundational reasoning capability remains unaffected. Additionally, since stable diffusion and LLM-grounded diffusion can only generate image results, we compare their performance in Table~\ref{table:comparision-with-stable-diffusion}.

\subsection{Qualitative Results}

\begin{figure}[t!]
    \centering
    \includegraphics[width=0.85\linewidth]{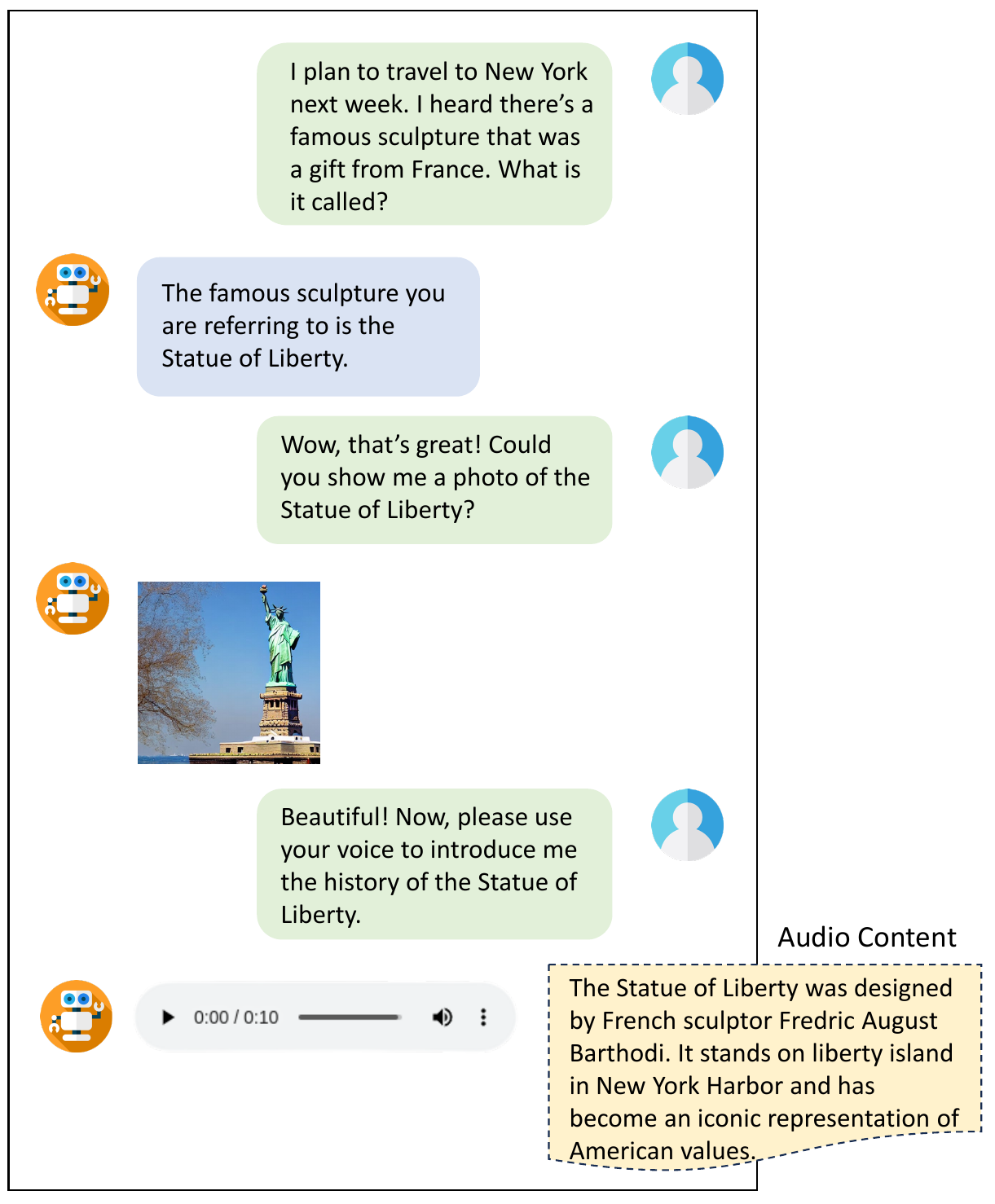}
    \caption{The proposed \methodname\ is able to generate non-text responses to fulfill the given instructions.}
    \label{fig:demo}
\end{figure}

In Figure~\ref{fig:Qualitative-results}, we provide a qualitative comparison of various methods based on instructions intended for image output generation. The results demonstrate that our proposed method aligns well with modality conversion models, yielding enhanced performance. Other methods, such as HuggingGPT, tend to use the instruction itself as the prompt to input into the modality conversion model, resulting in unsatisfactory outputs. Meanwhile, Figure~\ref{fig:demo} presents a demo (to be released) showcasing an example conversation between our model and users on the topic `The Statue of Liberty'. The interaction highlights our model's capability to discern the underlying intentions of the instructions, despite these conversational inputs being quite different from the instructions used during training. Consequently, it is capable of producing appropriate responses involving different modalities. See more results in the appendix.

\section{Conclusion}

In this paper, we have presented modality-aligned instruction tuning, a method designed to adapt text-based LLMs for generating non-text responses. This is the inaugural effort to align the outputs of LLMs and the inputs of modality conversion models from a data-driven perspective. A key advantage is that it allows LLMs to be trained without direct exposure to non-text modality data. Instead of delving into the complex real of multi-modal pre-training, they simply undergo a standard instruction tuning process, ensuring computational efficiency. To objectively evaluate our method, we have introduced a validation set comprising instructions that request text, image, and audio outputs. Based on this dataset, we examined several state-of-the-art LLMs in a few-shot setting and benchmarked our approach against them. We anticipate that our proposed \methodname\ will serve as a baseline for future studies in multimodal output LLMs. In future research, one promising direction could be to design adapters at the encoder stage that would enable the integration of multi-modal inputs. By seamlessly connecting this feature with the method proposed in this study, it could potentially give rise to a comprehensive Any-to-Any LLM capable of handling a wider array of tasks and functionalities.

\newpage
\appendix
\section*{Appendix}
\section{Instruction Generation}

Here, we show more details of the generation of modality-aligned instructions.

\subsection{Seed Instruction and Generation Prompts}

We have manually crafted 100 seed instructions, of which 50 are intended for image output and 50 for audio output. Listing~\ref{lst:image_seeds} and Listing~\ref{lst:audio_seeds} show some examples of image and audio seeds, respectively.

\begin{lstlisting}[language=json, caption=Examples of Image Seed Instruction, label=lst:image_seeds]
{
    "instruction": "Show me a good design for web UI.",
    "output": 
    {
        "type": "image",
        "response": "A master web UI design concept sketch featuring clean layout, intuitive navigation bar, and coherent color scheme."
    }
},
{
    "instruction": "Showcasing the city a hundred years from now.",
    "output": 
    {
        "type": "image",
        "response": "A futuristic cityscape with towering skyscrapers and flying vehicles."
    }
},
{
    "instruction": "Turn this sentence into an image: a photo of an astronaut riding a horse on Mars.",
    "output":
    {
        "type": "image",
        "response": "A photo of an astronaut riding a horse on Mars"
    }
}
\end{lstlisting}

\newpage
\begin{lstlisting}[language=json, caption=Examples of Speech Seed Instruction, label=lst:audio_seeds]
{
    "instruction": "Generate a voice clip to introduce Wuhan.",
    "output": 
    {
        "type": "audio",
        "response": "Wuhan, located in Central China, is known for its educational institutions like Wuhan University and landmarks such as the Yellow Crane Tower. Famous for its street food, the city embodies the spirit of modern China with its rich history and vibrant culture."
    }
},
{
    "instruction": "How do you pronounce this name: John?",
    "output": 
    {
        "type": "audio",
        "response": "John"
    }
},
{
    "instruction": "Perform a famous line from the movie 'The Godfather' using voice.",
    "output": 
    {
        "type": "audio",
        "response": "I'm going to make him an offer he can't refuse."
    }
},
{
    "instruction": "Read this tongue twister 'She sells sea shells by the seashore. The shells she sells are surely seashells. So if she sells shells on the seashore, I'm sure she sells seashore shells.'",
    "output": 
    {
        "type": "audio",
        "response": "She sells sea shells by the seashore. The shells she sells are surely seashells. So if she sells shells on the seashore, I'm sure she sells seashore shells."
    }
}
\end{lstlisting}

\newpage

After constructing seed instructions, we devised templates to prompt ChatGPT to generate instructions in bulk.

\begin{prompt}
You are asked to generate instructions based on given image descriptions. Here are some requests:\newline

1. Each instruction should be short and concise, ideally 1 to 2 sentences long in English.

2. Each instruction should be based on the image descriptions. But, try not to directly use the contents in descriptions to form the instruction, and reduce the use of the words that already appeared in descriptions.

3. Do not generate a request that might lead to confusing answers. For example, do not ask to generate a character without any further information.

4. The language of instructions should be diverse, try to use different verbs for each instruction to maximize the diversity. Either an imperative sentence or a question is permitted.\newline

Here are some examples:

\textcolor{commentgreen}{// Each time we randomly insert 3 examples from seed instructions here}\newline

[A photo of avocado]

Instruction: Visualize the answer to this riddle: I'm a fruit that's green when I'm raw and black when I'm ripe. I'm buttery but I'm not a dairy product. What am I?

[A futuristic cityscape with towering skyscrapers and flying vehicles.]

Instruction: Showcasing the city a hundred years from now.

[Mythical creature with dragon's body, phoenix's wings, and both covered in flames.]

Instructions: Create a visual representation of a mythical creature that combines elements of a dragon and a phoenix. \\

The following are some image descriptions, Please generate an appropriate instruction for each of them:

\textcolor{commentgreen}{// Image captions are sampled from LAION-aesthetics}\\

[A large passenger airplane flying over some palm trees]

[A dog watching a man cut a piece of food]

...
\label{prompt:image-instruction-generation}
\end{prompt}

The template provided above illustrates the prompt input for OpenAI's GPT API~\cite{openai2023chatgpt}, tailored for generating image instructions. To generate speech instructions, we simply adjust some keywords, such as changing `image' to `speech', and use the text-to-speech training set as a reference. For each time the API is utilized, three examples from the seed instructions are embedded into the prompt, and 60 captions are sampled from the reference dataset. Consequently, the API can deliver up to 60 instructions per call, with a cost of $<$ USD\$ 0.01 each time.

\begin{figure}[t!]
\centering
    \begin{subfigure}{0.45\textwidth}
    \centering
        \includegraphics[width=0.8\linewidth]{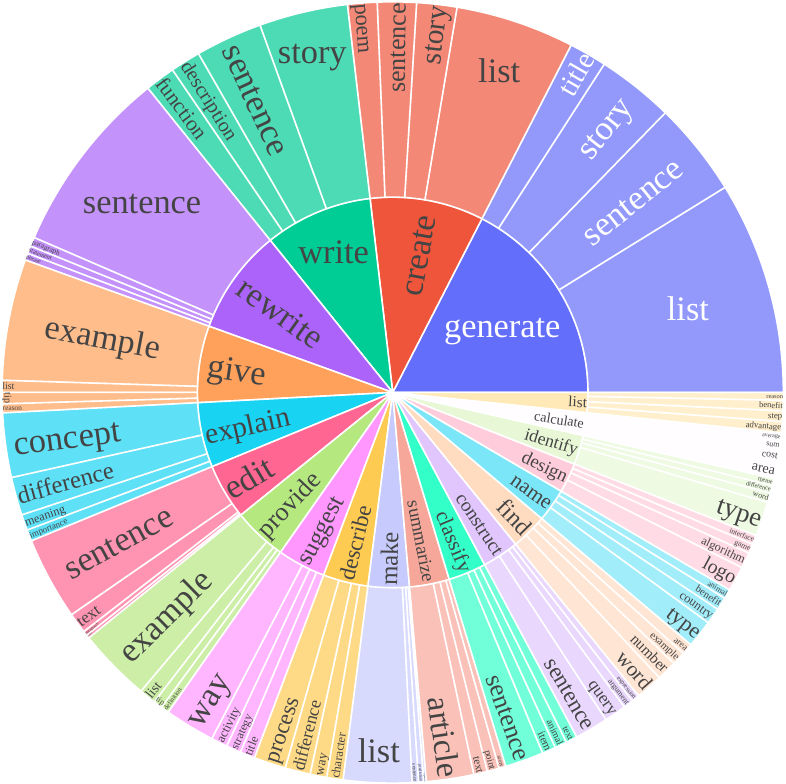}
        \caption{Text-only Instructions~\cite{alpaca}}
        \label{fig:subplot1}
    \end{subfigure}
    \begin{subfigure}{0.45\textwidth}
    \centering
        \includegraphics[width=0.8\linewidth]{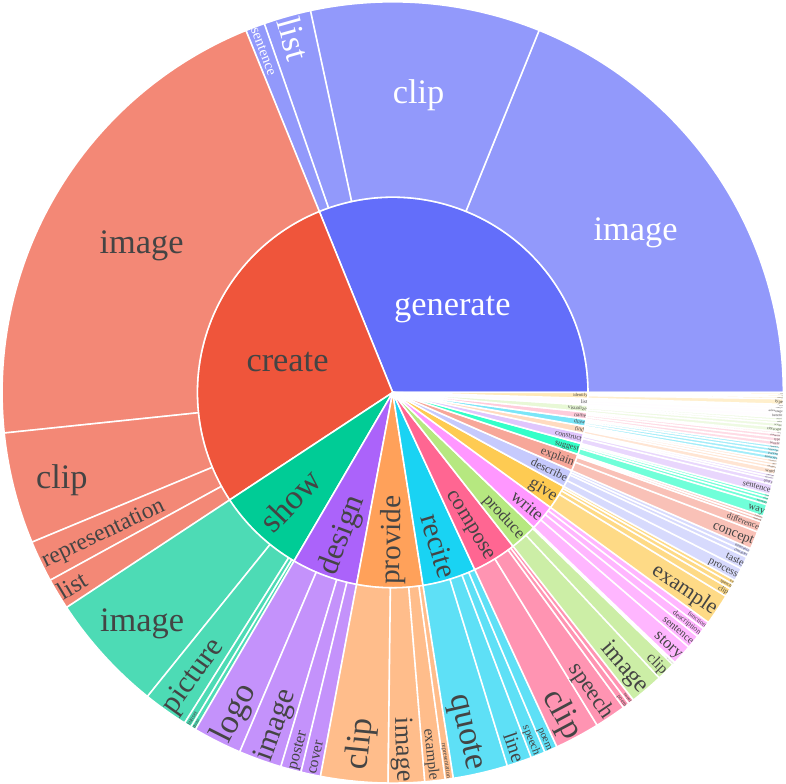}
        \caption{Modality-aligned Instructions (Ours)}
        \label{fig:subplot2}
    \end{subfigure}
    \caption{Comparison of the distribution of some top root verbs and corresponding nouns between the text-only instructions generated by Alpaca~\cite{alpaca} and ours modality-aligned instructions.}
    \label{fig:Comparison-instruction-distribution}
\end{figure}

Based on the above method, we generated approximately 40k instructions with expected image or audio outputs. Subsequently, we removed those that were malformed and could not be parsed. Additionally, we eliminated instructions that requested nonspeech audio, such as the sounds of a cat purring, thunder clapping, or music, as well as those asking for sentences to be read in other languages, such as Japanese or Arabic. 

Figure~\ref{fig:Comparison-instruction-distribution} contrasts the distribution of root verbs and their corresponding nouns between our modality-aligned instructions and a text-based instruction dataset produced by Alpaca~\cite{alpaca}. As depicted, our instructions contain a substantial number of non-text directives, like ``Generate image", ``Design logo", and ``Recite quote", which were deliberately excluded by Alpaca. We merge 

\section{Evaluation}

\subsection{Few-shot Inference for Text-based LLMs}

In the main paper, we have explained that with proper prompts, the pre-trained pure text-based LLMs are also capable of analyzing the intended output modality of given instructions. Therefore, we designed the following template prompt to evaluate these LLMs.

\begin{prompt}
You are a helpful assistant. Given an instruction, your task is to generate an appropriate JSON-formatted response. Here's how you can do it:\newline

1. Determine the expected output modality of the given instruction. The available options are `text', `image', and `audio'.

2. Place the expected modality in the `type' field of your response.

3. Proceed to fulfill the instruction based on the determined modality:

\begin{itemize}
    \item If the desired modality is `text' or `speech', provide the response directly in the `response' field.
    \item If the desired modality is `image', generate a concise and brief prompt. This prompt will be sent to a text-to-image model, which will then generate the image that fulfills the request.
\end{itemize}

Instruction: ``Show me the architectural style of the Baroque era."

Response: `{``type": ``image", ``response": ``Ornate, grand Baroque-style palace with dramatic lighting and complex shapes."}'\newline

Instruction: ``Show me a tranquil Japanese Zen garden."

Response: '{``type": ``image", ``response": ``A tranquil Japanese Zen garden with raked sand, smooth rocks, moss, and a wooden bridge over a small pond."}'\newline

Instruction: ``How do you say `October 1, 2000' in spoken English?"

Response: `{``type": "audio", ``response": ``October first, two thousand."}'\newline

Instruction: ``Create an audio clip with Easter as the theme."

Response: `{``type": ``audio", ``response": ``Easter, a cherished holiday, is celebrated as a time of renewal and rebirth. It brings joy with its symbols of spring, such as eggs, bunnies, and flowers."}'\newline

Instruction: ``What are the three primary colors?"

Response: `{``type": ``text", ``response": ``The three primary colors are red, blue, and yellow."}'\newline

Instruction: \textbf{\{instruction\}}

Response: 
\end{prompt}

During the inference stage, the `\textbf{\{instruction\}}' will be replaced with actual inputs. While this method can measure the performance of existing methods to a certain extent, it's important to note that the outputs can vary significantly depending on the method employed. A clear drawback of such approaches is the need to encode lengthy explanations and examples into the prompt each time the LLM is tasked with generating a response. This can be inefficient; for instance, the template prompt itself contains over 400 tokens, leading to significant redundancy. To ensure a fair comparison, all text-based LLMs undergo few-shot inference using the identical template.

\subsection{Evaluation Instructions}

There is no off-the-shelf suite to evaluate multi-modality output LLMs. Given that the training set consists of short instructions generated by GPT, employing similar instructions as a validation set for assessing model performance is inappropriate. Hence, rather than using short instructions, we utilize lengthier, conversation-like inputs for validation. This approach avoids potential overlap with the training set and better assesses the model's generalization ability. We manually gathered conversation segments as validation inputs (200 for each modality, excluding text). These segments, combined with our instruction generation pipeline and manually filtered, yielded a total of 800 validation instructions per modality. For the text modality, given the abundance of existing evaluation sets, we directly sampled 800 entries from datasets like TruthfulQA~\cite{lin2021truthfulqa} and MMLU~\cite{hendrycks2020measuring}. Listing~\ref{lst:validation_instruction} shows some examples from the validation set.

\begin{lstlisting}[language=json, caption=Examples of Validation Instruction, label=lst:validation_instruction]
{
"instruction": "I am confused by so many types of camelids. Could you show me a photo of different types of these creatures? Thank you!",
"output": {
    "type": "image",
    "response": "A photo compares a llama, an alpaca, a vicuna, and a Gunaco side by side.",
    "image_id": "24531"
}},{
"instruction": "Can you show me the famous Japanese painting which includes wave and mountain fuji?",
"output": {
    "type": "image",
    "response": "The Great Wave off Kanagawa.",
    "image_id": "42153"
}}
\end{lstlisting}

\begin{figure*}[t!]
    \centering
    \includegraphics[width=\linewidth]{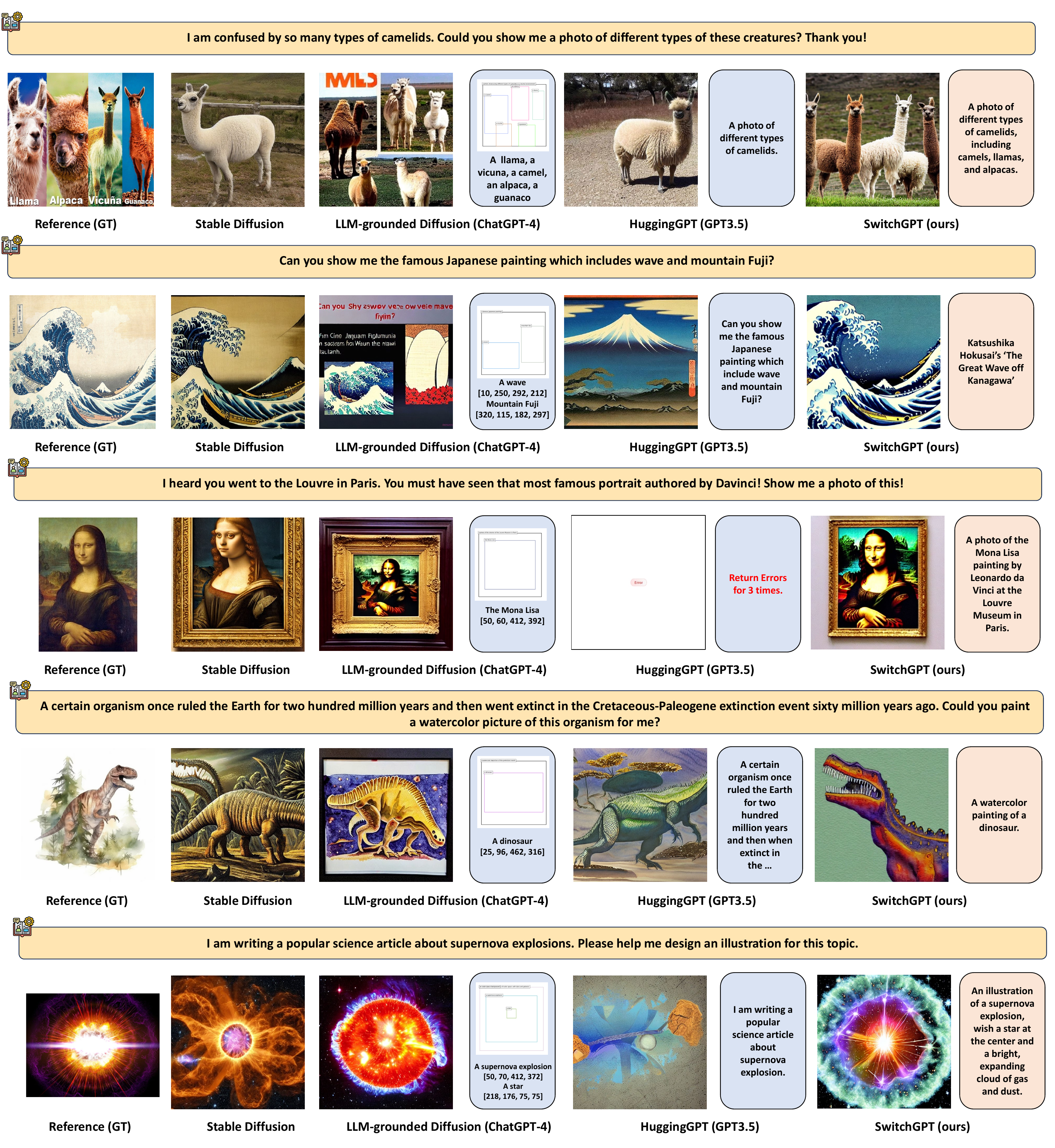}
    \caption{Qualitative Results (Best viewed zoomed in).}
    \label{fig:supp-qualitative-results}
\end{figure*}

\begin{figure*}[t!]
    \centering
    \includegraphics[width=\linewidth]{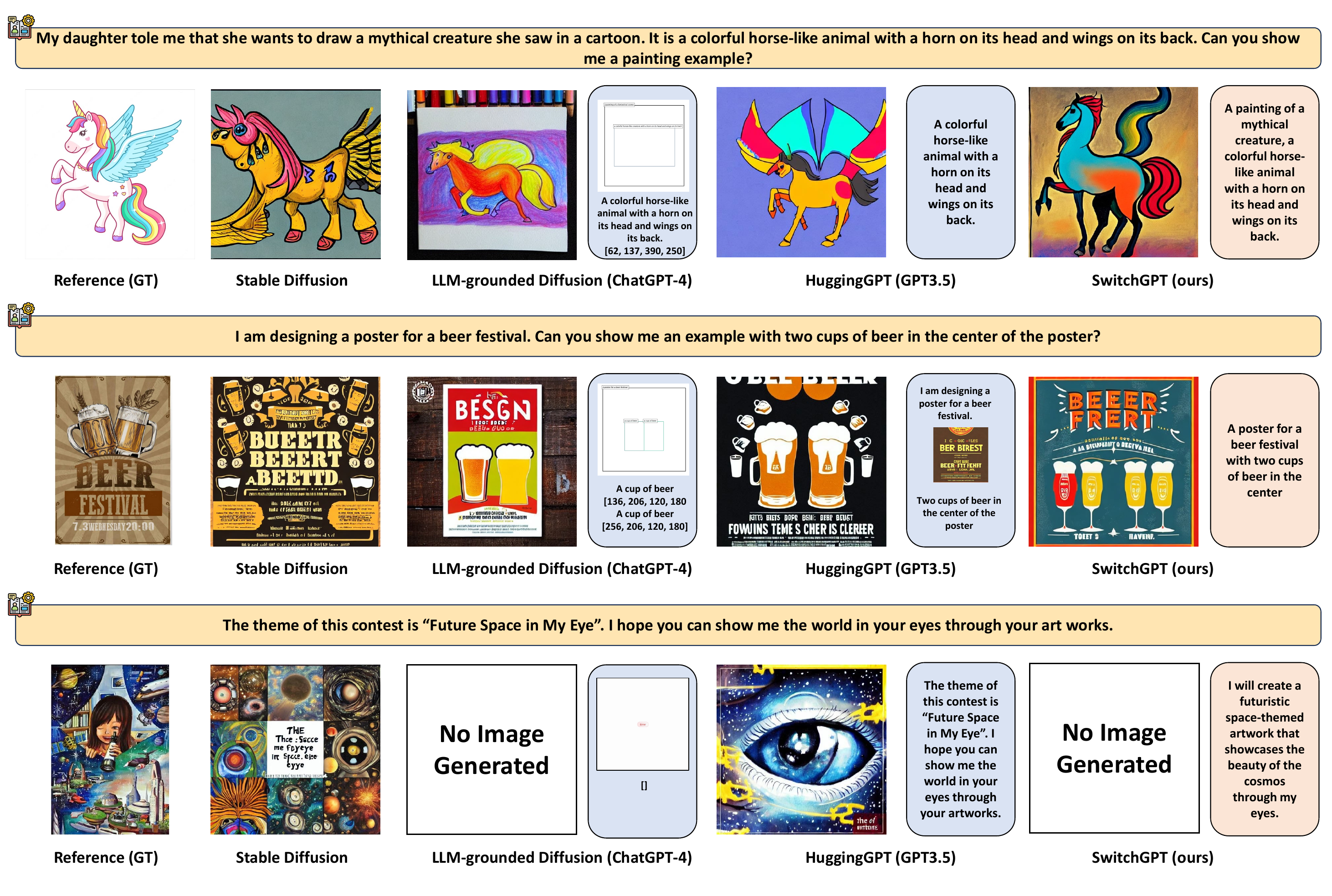}
    \caption{Failure Cases.}
    \label{fig:failures}
\end{figure*}

\subsection{Metrics}

\textbf{Modality Classification Accuracy.} We employ classification accuracy as the metric to evaluate whether a method can accurately interpret the desired output modality specified by the given instructions. It can be simply calculated as follows:

\begin{equation}
   A = \frac{n_{correct}}{n_{total}}
\end{equation}

\noindent In this equation, $n_{correct}$ represents the number of correctly classified modalities, while $n_{total}$ denotes the total count of test instructions.

\noindent\textbf{Language.} For language evaluation, existing datasets resembling multiple-choice formats already provide several correct and incorrect answers for each instruction. We employ the BLEU score to match the responses generated by LLMs against the reference answers. Subsequently, we select the response that is most similar to determine the final accuracy.

\noindent\textbf{Image.} We utilize two widely accepted metrics to assess the quality of generated images based on their instructions: the CLIP and FID scores. The CLIP score measures the relevance between the generated image and the corresponding ground-truth descriptions. For instance, Listing~\ref{lst:validation_instruction} presents several records from the test set. Within each record, there's a textual description of the anticipated image in the `response' field, which we employ to compute the CLIP score. On the other hand, the FID score evaluates the quality of image generation by calculating the Fréchet inception distance between the generated images and their reference counterparts. The referenced image is indicated by the `image\_id' field in the records.

\noindent\textbf{Speech.} All LLMs utilize the same speech generation model, resulting in nearly the same audio quality. Therefore, our primary concern is whether the content aligns with the instructions. As such, we approach the evaluation of this task as a translation/generation task, employing the BLEU score to measure the similarity between the response and the ground truth.

\section{Qualitative Results}

In the main paper, we have presented some qualitative results. Here, we show more comparisons between our methods and existing methods.

\subsection{More comparisons}

Figure~\ref{fig:supp-qualitative-results} displays additional examples from both existing methods and our approach. Several interesting observations are as follows:

\begin{itemize}
    \item While Stable Diffusion~\cite{rombach2022high} was not specifically optimized for processing extended conversation-like inputs, it remains adept at capturing key terms within such instructions. Thus, if the anticipated output is evident or directly associated with the input, Stable Diffusion can still produce satisfactory results. For instance, in the last row of Figure~\ref{fig:supp-qualitative-results}, the term `supernova explosions' is present in the input instruction, prompting Stable Diffusion to generate a commendable representation of this concept. Similarly, in the second row of Figure~\ref{fig:supp-qualitative-results}, even though the instruction does not explicitly mention the painting's title — `The Great Wave off Kanagawa' — Stable Diffusion creates an almost perfect image (though the position and size of the mountain are different from the referenced image) based on the given input. This proficiency might be attributed to Stable Diffusion's prior exposure to similar images during its training. Phrases like `wave', `Japanese painting', and `Mount Fuji' are sufficient cues for it to recognize and retrieve from its stored knowledge. In essence, for widely recognized concepts, Stable Diffusion can produce commendable results with sparse information, leveraging its extensive pre-training. However, lesser-known concepts or requests that require stronger reasoning ability, might not always yield promising results.
    \item LLM-grounded Diffusion~\cite{lian2023llm}, leveraging the extremely strong reasoning capabilities of ChatGPT-4, yields promising results. However, when converted to ChatGPT-3.5, it fails to produce any results for a majority of inputs. Notably, for concepts that are composed of multiple sub-items, the LLM-grounded Diffusion creates layouts for each individual item. This approach can inadvertently segment a cohesive concept into disjointed fragments. For instance, as depicted in the second row of Figure~\ref{fig:supp-qualitative-results}, the LLM-grounded Diffusion separated the painting into two distinct elements: a wave and Mount Fuji. Consequently, the final output appears as if two unrelated images were pieced together, leading to gaps and extraneous content in their juxtaposition.
    \item In the case of HuggingGPT~\cite{shen2023hugginggpt}, given its support for multi-step tasks, it occasionally produces unnecessary and irrelevant modality outputs. For instance, it might frequently provide an audio clip alongside the output image. Additionally, HuggingGPT sometimes struggles to properly parse its own output, leading to unexpected errors. As depicted in the third row of Figure~\ref{fig:supp-qualitative-results}, HuggingGPT failed thrice in executing the provided instruction. Another limitation is its misalignment between its output and the input of the text-to-image models. Consequently, it often returns a segment of the instruction as a prompt for image generation. This results in its outputs being notably similar to the original stable diffusion in many cases.
\end{itemize}

\subsection{Failure Cases}

In Figure~\ref{fig:failures}, we present several failure cases of our proposed model, \methodname.

In the first row, all methods fail to recognize the intended entity "a colorful unicorn" from the inputs. Instead, they default to using the description ``a colorful horse-like animal" provided in the given instruction, leading to erroneous outputs.

The second row illustrates that when given an instruction requesting a specific layout, such as ``two cups of beer in the center", our method predominantly relies on the modality conversion models to interpret the prompt, even though it offers a relatively accurate description. On the other hand, LLM-grounded Diffusion excels in this instance, being optimized for such scenarios. It directly generates a layout with bounding boxes for two beers, yielding an optimal result. Interestingly, HuggingGPT segments the instruction into two parts. It uses the first segment ``I am designing a poster for a beer festival" for a text$\rightarrow$image task and then employs the generated image, along with the second part ``two cups of beer in the center of the poster", as prompts for an image$\rightarrow$image task. The final output successfully places ``two cups of beer in the center."

In the third row, both LLM-grounded Diffusion and \methodname\ fail to produce an image, albeit for distinct reasons. The former's failure arises because ChatGPT-4 returns an empty object list for this instruction, preventing LLM-grounded Diffusion from creating any images. In contrast, our model struggles to discern the intended modality of the given instruction, leading it to produce a purely textual response.

\end{document}